\documentclass{article}

\usepackage{microtype}
\usepackage{graphicx}
\usepackage{booktabs}
\usepackage{hyperref}
\usepackage{amsmath}
\usepackage{amssymb}
\usepackage{mathtools}
\usepackage{amsthm}
\usepackage[normalem]{ulem}
\usepackage[capitalize,noabbrev]{cleveref}
\usepackage{enumitem}

\usepackage[accepted]{icml2026}

\usepackage{subcaption} 
\usepackage{enumitem}

\newcommand{\fim}{FIM}
\newcommand{\ltr}{LTR}
\newcommand{\pz}{p_z}

\definecolor{ltrblue}{HTML}{CCE3F0}   
\definecolor{fimpink}{HTML}{F5E4ED}     
\definecolor{finewebgreen}{HTML}{CCECE3} 

\definecolor{prefixcolor}{HTML}{8A3A0A}
\definecolor{suffixcolor}{HTML}{7A1F55}

\icmltitlerunning{Memorization Dynamics of Fill-in-the-Middle Pretraining}

\begin{document}

\twocolumn[
  \icmltitle{Memorization Dynamics of Fill-in-the-Middle Pretraining}

  \begin{icmlauthorlist}
    \icmlauthor{Tobias von Arx*}{eth}
    \icmlauthor{Tanguy Dieudonné*}{eth}
  \end{icmlauthorlist}

  \icmlaffiliation{eth}{Department of Computer Science, ETH Zurich, Zurich, Switzerland}

  \icmlcorrespondingauthor{Tobias von Arx}{tvonarx@ethz.ch}
  \icmlcorrespondingauthor{Tanguy Dieudonné}{tdieudonne@ethz.ch}  

  \icmlkeywords{memorization, fill-in-the-middle, large language models, pretraining}

  \vskip 0.3in
]

\printAffiliationsAndNotice{\icmlEqualContribution}

\begin{abstract}
Fill-in-the-middle (\fim{}) is a pretraining objective widely used to equip causal language models with infilling ability, yet its effect on verbatim memorization remains underexplored.
We study the memorization dynamics of \fim{} in a controlled setting by pretraining matched Llama 3.2 models with FIM and standard left-to-right (\ltr{}) objectives on a FineWeb-Gutenberg corpus containing repeated Gutenberg excerpts.
With prefix-based probes, \fim{} more often recovers short or partially matching spans, while \ltr{} more often assigns high confidence to long exact continuations.
We observe that verbatim extraction under \fim{}-training grows approximately linearly with repetitions over the tested range.
Evaluating native \fim{}-format probes reveals that suffix context is not sufficient: verbatim recall under \fim{}-training remains strongly anchored in prefix context.
Our results also show that evaluating only one span length or probing format can miss important nuances in memorization behavior.

\end{abstract}

\section{Introduction}

Large language models can reproduce training data, including rare strings, private information, code, and book passages \citep{carlini2019secret,carlini2021extracting,nasr2025scalable,cooper2025extracting}. Early work measured unintended memorization with synthetic canaries and exposure scores \citep{carlini2019secret}; later attacks extracted real training examples \citep{carlini2021extracting,nasr2025scalable}. Recent work studies leakage beyond greedy decoding, including probabilistic extraction \citep{hayes2025measuring}, book-level extraction \citep{cooper2025extracting}, and membership-style tests \citep{mattern2023membership,shi2024detecting}.

Repetition is one of the clearest predictors of memorization. Deduplication reduces verbatim generations \citep{lee2022deduplicating}; duplicate count predicts regeneration \citep{kandpal2022deduplicating}; and controlled injections are recovered more often as exposure increases \citep{huang2024demystifying}. Attribution remains difficult because prior predictability, near duplicates, tokenization, prompt position, and available context can all affect recovery \citep{kharitonov2022bpe,zhang2023counterfactual,shilov2026mosaic,liu2024lost,xu2026positional}.

We study fill-in-the-middle (\fim{}), a common pretraining objective for causal language models \citep{bavarian2022efficient}. Standard left-to-right (\ltr{}) training predicts each token from its prefix. \fim{} training moves a target middle span after prefix and suffix, separated by sentinel tokens, such that during training, the target is exposed to right context as well as left context. Infilling is used in systems such as DeepSeek-v3, InCoder, StarCoder, and Code Llama \citep{deepseekai2025deepseekv3technicalreport,fried2023incoder,li2023starcoder,roziere2023code}. Prior work has mainly emphasized infilling utility; here we ask how the objective impacts verbatim extraction.

We conduct a controlled study comparing standard \ltr{} and \fim{} pretraining under matched architecture and data source, asking three related questions:

\begin{enumerate}[label=(\roman*)]
    \item How does \fim{} impact verbatim memorization across target span lengths, extraction thresholds, and repetition?
    \item Under native \fim{} prompting, how do prefix context, suffix context, and sentinel tokens contribute to verbatim memorization?
    \item Are the observed effects specific to extraction geometry, or explained by broad model-quality differences?
\end{enumerate}

\section{Study Design}
\label{sec:study-design}

We compare paired \ltr{} and \fim{} models trained on the same data, architecture and parameters. Our controlled conditions let us attribute differences in memorization to the pretraining format.

We release our code at \url{https://github.com/tobiasvonarx/memorization-study-fim}.

\subsection{Matched Training with Controlled Repetition}
\label{sec:matched-training}

The bulk corpus is FineWeb 100B, while our controlled memorization corpus consists of Project Gutenberg books \citep{penedo2024fineweb,gutenberg}. We score 4096-token windows of Gutenberg books with a Llama 3.2 model \citep{grattafiori2024llama3herdmodels} trained only on FineWeb, in order to filter out pre-memorized, outlier, and duplicate windows. The resulting cleaned set of excerpts is split into 12 repetition buckets of 2{,}810 excerpts with exposures from 1 to 128. We balance bucket assignment by prior perplexity.

We build two corpora from the same data sources. The \ltr{} corpus keeps autoregressive order. The \fim{} corpus rewrites examples into sentinel-delimited prefix--suffix--middle order, where the spans are randomly partitioned.
In particular, repeated \fim{} copies use different split points, so repetition is document-level exposure rather than fixed middle-span exposure.
The \fim{}-corpus contains \(50\%\) \fim{}-documents for FineWeb (the rest being \ltr{}) and \(100\%\) \fim{}-documents for Gutenberg.

Both models use an identical Llama 3.2 3B architecture and are trained over one epoch of \(\approx103\)B tokens (\(\approx95\%/5\%\) FineWeb/Gutenberg). Further experimental details are listed in \cref{app:experimental-details} and model size is ablated in \cref{app:ablation-model-size}.

\subsection{Downstream performance}
We evaluate both models on 8 tasks of the LM Evaluation Harness \citep{lm-eval-harness}, and observe that both models achieve nearly identical performance. Detailed metrics are provided in \cref{app:evaluation-results}. We conclude that differences in memorization are not due to differences in model capabilities in the context of our study.

\section{Prefix-only Extraction}
\label{sec:prefix-only-extraction}

We compare FIM and LTR with the same prefix-only probe: using 100 prefix tokens to predict a span $z$ of $M=32$ target tokens. For each repetition bucket, we probe both models on the same Gutenberg windows, sampling 10 disjunct windows per excerpt.

\begin{figure*}[t]
    \centering
    \includegraphics[width=.9\linewidth]{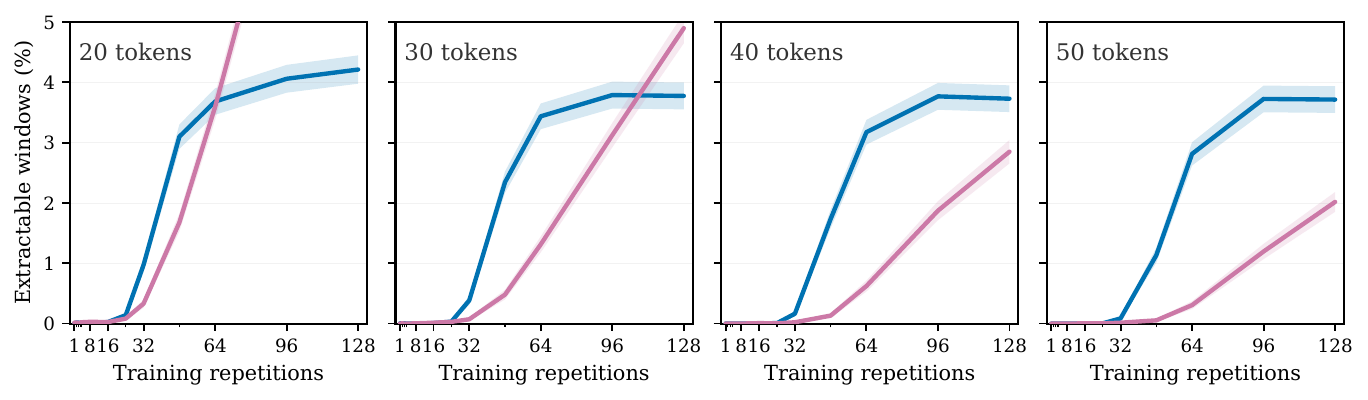}
    \caption{
    \textbf{Extraction rates under varying target lengths} show that the repetitions required for \colorbox{fimpink}{\fim{}} to overtake \colorbox{ltrblue}{\ltr{}} increases with span length, because longer spans favor \ltr{}'s heavier tail. Curves show the fraction of windows with \(p_z \geq 0.1\%\) for the first 20, 30, 40, and 50 target tokens; all panels use the same y-axis scale. Shaded bands denote nominal 95\% confidence intervals for the per-window rate.
    }
    \label{fig:ltr-cooper-span-extractability}
\end{figure*}

We report two criteria.
First, inspired by \citet{cooper2025extracting}, exact extraction computes \(p_z=\prod_{i=1}^M q_i\), where \(q_i\) is the top-\(k\)-renormalized probability of the $i$-th target token under \(k=40,T=1\).
A target is called \textit{extractable} if \(p_z \ge 0.1\%\).
Second, we generate \(M\) tokens starting from the prefix autoregressively and report ROUGE-L \citep{rouge}, with ROUGE-L \(\ge 0.5\) indicating \textit{high-overlap recovery} following \citet{chen2025parapo}.
Using $M=32$ lets us evaluate both criteria on the same windows.
This is less strict per token than the $M=50$ setting in \citet{cooper2025extracting} ($80.6\%$ vs. $87.1\%$ geometric mean). We vary \(M\) in \cref{fig:ltr-cooper-span-extractability}.

\begin{figure}[t]
    \centering
    \begin{subfigure}[t]{0.48\linewidth}
        \centering
        \includegraphics[width=\linewidth]{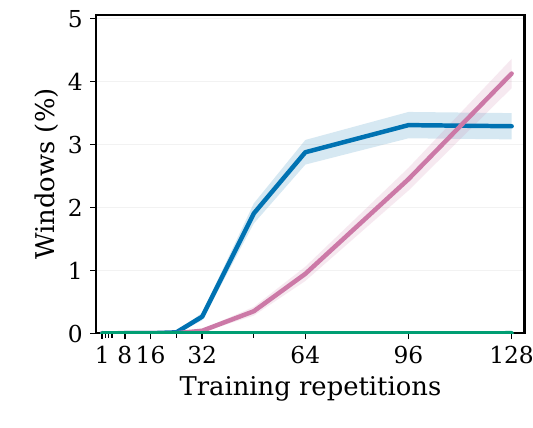}
        \caption{Verbatim extraction rate}
        \label{fig:ltr-memorization-rates-cooper}
    \end{subfigure}
    \hfill
    \begin{subfigure}[t]{0.48\linewidth}
        \centering
        \includegraphics[width=\linewidth]{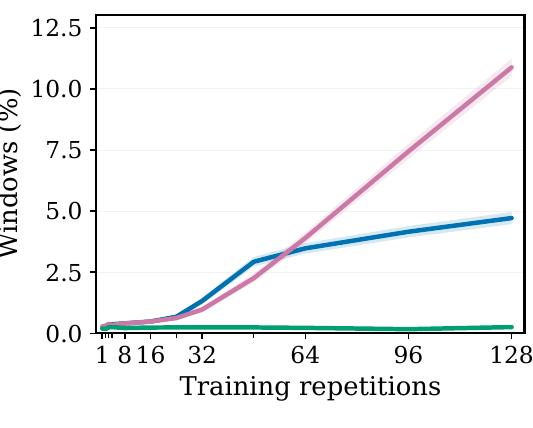}
        \caption{High-overlap recovery rate}
        \label{fig:ltr-memorization-rates-rouge}
    \end{subfigure}
    \caption{
    \textbf{Memorization across repetition buckets.} For strict full-span extraction, \colorbox{ltrblue}{\ltr{}} is higher in aggregate, but \colorbox{fimpink}{\fim{}} extracts more windows at the largest repetition bucket. \colorbox{fimpink}{\fim{}} yields stronger high-overlap recovery for high repetitions. \colorbox{finewebgreen}{FineWeb} is the baseline trained only on FineWeb. Shaded bands denote nominal 95\% confidence intervals for the per-window rate.
    }
    \label{fig:ltr-memorization-rates}
\end{figure}

For the exact extraction criterion, \ltr{} overall memorizes more windows: 3,279 windows satisfy \(p_z \geq 0.1\%\), versus 2,230 for \fim{}.
\fim{} is slightly higher on broader recovery measures, including mean ROUGE-L (0.198 for \fim{} vs 0.190 for \ltr{}), and mean top-$k$ support rate (87.09\% vs 86.18\%), i.e., the fraction of reference tokens contained in the top-$k$ of logits with \(k=40\).
The low memorization rate is partly due to probe position. Beginning-of-excerpt probes memorize significantly more than randomly sampled windows (\cref{fig:ltr-window-position-rouge-1b-3b} of \cref{app:additional-figures}).

\begin{figure}[h]
    \vspace{1em}
    \centering
    \includegraphics[width=.9\linewidth]{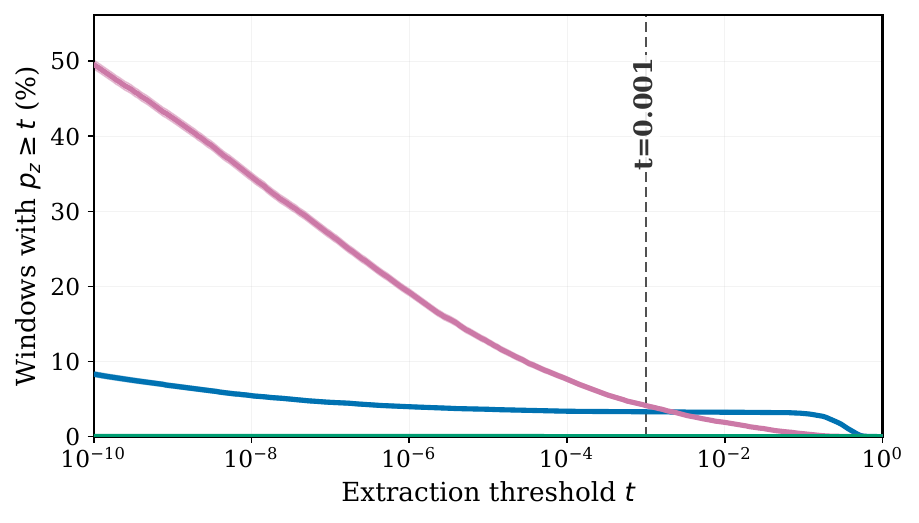}
    \caption{
    \textbf{Extraction survival curves} at repetition \(128\) show that \colorbox{fimpink}{\fim{}} assigns more mass to moderately likely targets, but \colorbox{ltrblue}{\ltr{}} has the heavier high-confidence tail. Each line gives the percentage of evaluated target windows with \(p_z \geq t\) as the extraction threshold \(t\) varies. The 95\% confidence intervals are smaller than the line width.
    }
    \label{fig:ltr-cooper-pz-survival}
\end{figure}

While the \fim{} model's support is higher, probability mass is less concentrated on complete 32-token continuations. The exact extraction criterion is strict, such that few low-probability tokens can collapse the \(\pz{}\) of a target span.
A threshold sweep at repetition \(128\) confirms this: \cref{fig:ltr-cooper-pz-survival} shows that \fim{} has more mass at moderate $\pz{}$, but \ltr{} has the heavier tail, and therefore extracts more at the \(0.1\%\) threshold.

In line with \citet{huang2024demystifying}, we find that non-trivial repetitions are required for memorization.
This is expected, especially at the 3B model scale, since memorization increases with model capacity \citep{carlini2023quantifying}.
We study a 1B ablation in \cref{app:ablation-model-size}. 
With more repetitions, \ltr{} extraction shows diminishing returns, consistent with the logarithmic trend reported in \citet{carlini2023quantifying}.
While \fim{}-extraction rises more steadily with repetitions, it remains low for small repetition counts.
We ablate the target length in \cref{fig:ltr-cooper-span-extractability} and conclude that the number of repetitions required for \fim{} to surpass \ltr{} in extraction increases with span length. This is because a longer target makes extraction stricter, such that \ltr{}'s heavy-tailed distribution dominates.

We analyze attention patterns to further contextualize our insights. For each target-position prediction query, we partition the attention between (i) the prefix tokens and (ii) the already-seen target tokens. The latter is zero for the first target token of the target span and, for later positions, includes all earlier target tokens in the target span. We average over target positions and windows and report the mean attention allocation in \cref{tab:ltr-attention-partition-ci}. The \fim{} model places more attention on the prefix and less on already-seen target tokens compared to the \ltr{} model. 

Our observations can be explained by the structure of the \fim{} objective.
Repeated \ltr{} examples present each passage under the same left-to-right view.
This concentrates probability mass into fewer long continuations, leading to the heavy-tailed distribution with increased extraction.
Repeated \fim{} examples instead expose the same passage through varied prefix--middle--suffix decompositions, spreading mass across more partial reconstructions and broadening recoverability.

\begin{table}
  \centering
  \caption{Mean attention allocation during prediction of the target span. Both models rely primarily on the prefix, but \fim{} relies on it more strongly, while \ltr{} allocates relatively more attention to earlier target tokens. Nominal 95\% confidence intervals are below \(10^{-4}\).}  
  \label{tab:ltr-attention-partition-ci}
  \begin{tabular}{lrr}
    \toprule
    Model & Prefix attention & Previous-target attention \\
    \midrule
    \ltr{} & $0.604$ & $\mathbf{0.396}$ \\
    \fim{} & $\mathbf{0.646}$ & $0.354$ \\
    \bottomrule
  \end{tabular}
\end{table}

\vspace{-1.25mm}
\section{Native \fim{} probing}
\label{sec:native-fim-probing}

Since the native \fim{}-format includes both left and right context, it fundamentally differs from the prefix-only extraction prompt.
We study the \fim{}-native format to evaluate how prefix and suffix context redistribute attention and contribute to memorization.
As before, we sample 10 disjunct windows for each excerpt and the target remains 32 tokens.
However, the 100-token context is now split across prefix and suffix.
Additionally, we focus our analyses on the 128-repetition bucket, in which memorization is most prevalent.
Note that this probing format includes the \fim{}-sentinel tokens, so even if the suffix is empty, it still differs from the prefix prompt evaluated in \cref{sec:prefix-only-extraction}. 

\begin{figure}
    \centering
    \includegraphics[width=.95\linewidth]{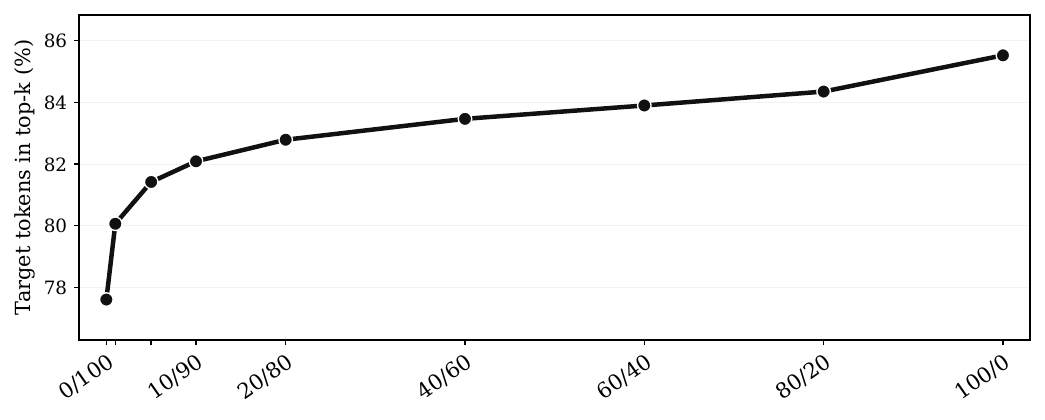}
        \caption{\textbf{Target-token top-$k$ support} under native FIM geometry at 128 repetitions shows that memorization improves monotonically as more of the 100-token context budget is allocated to the prefix rather than the suffix. The x-axis varies \texttt{prefix}/\texttt{suffix} lengths. The line shows the percentage of target tokens included in top-$40$ support. The 95\% confidence intervals are smaller than the line width.}
    \label{fig:top-k-native}
    \vspace{-1em}
\end{figure}

In \cref{fig:top-k-native}, we vary the prefix--suffix split around a fixed target to test which side of the native \fim{} context contributes more to memorization support.
As the prefix grows and the suffix shrinks, top-\(k\) support increases monotonically.
The same trend holds within all repetition buckets and for both extraction rates and target likelihood (see \cref{app:additional-figures}).
In all repetition buckets, moving from suffix-only to prefix-only context, target perplexity falls from 60.23 to 27.93, while top-\(k\) support rises from 77.60\% to 85.52\%.
The sharp drop when little or no prefix is available reflects the autoregressive structure of causal language models: without left context, the model has no reliable starting point for generating the middle span.

\begin{figure}[h]
  \centering
  \includegraphics[width=.95\linewidth]{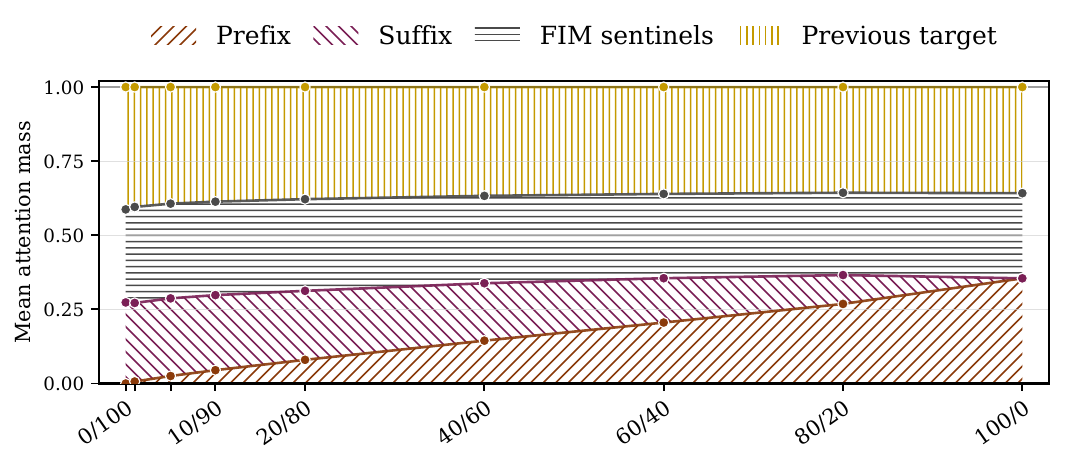}
  \caption{\textbf{Attention allocation} under native \fim{} probing shows that the model uses both surrounding contexts, with more attention on the prefix than the suffix, and shifts attention toward earlier target tokens when little prefix is available. The stacked areas show mean attention mass assigned to prefix tokens, suffix tokens, \fim{} sentinels, and earlier target tokens within the target span, averaged over target-token prediction queries and repetition buckets. The x-axis varies \texttt{prefix}/\texttt{suffix} lengths.}
  \label{fig:native-fim-attention-stack}
\end{figure}

While prefix-heavy native \fim{} prompts elicit stronger memorization, the suffix still provides conditioning. The attention analysis in \cref{fig:native-fim-attention-stack} shows substantial attention allocated to both prefix and suffix, with the prefix receiving slightly more attention. For prompts with very little prefix, the model compensates by attending more heavily to preceding tokens of the target span.

\begin{figure}[t]
  \centering
  \includegraphics[width=\linewidth]{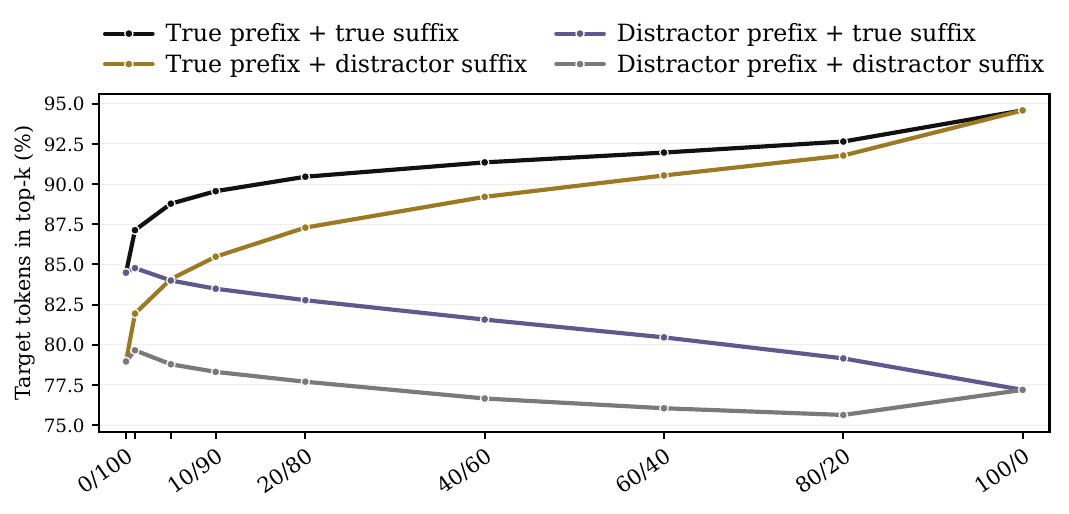}
  \caption{
  \textbf{Target-token top-$k$ support} under native \fim{} prompting at 128 repetitions and different distractor conditions. Replacing the prefix harms recall more than replacing the suffix, confirming that prefix context is the stronger driver of memorization.
  The x-axis varies \texttt{prefix}/\texttt{suffix} lengths. The 95\% confidence intervals are smaller than the line width.}
  \label{fig:prefix-rescue-topk}
  \vspace{-1mm}
\end{figure}

To isolate the contribution of prefix and suffix context directly, we keep the target fixed and replace the prefix, the suffix, or both with same-length unrelated \textit{distractor spans} from different Gutenberg excerpts.
We consider excerpts in the 128-repetition bucket and vary the prefix--suffix ratio, keeping the total context budget fixed.
\cref{fig:prefix-rescue-topk} shows the top-$k$ support in this setting.
We deduce that prefix and suffix are not equally significant.
Recall is strongest when the available context is allocated to the prefix.
As expected, the full prompt yields the strongest top-\(k\) support across the sweep, serving as an upper-bound reference for the distractor-span conditions.
While replacing the suffix with a distractor reduces recall, replacing the prefix has a significantly larger effect.
When both sides are replaced by distractors, we verify that support drops sharply, confirming that the effect is not only due to prompt length or sentinel structure.

\section{Conclusion}

Matched \ltr{} and \fim{} models trained on a corpus containing repeated book excerpts show that the pretraining objective shapes how memorization accumulates. Under prefix-only probes, \fim{} improves short-span and overlap-based recovery, especially at high repetitions, while \ltr{} produces more high-confidence long exact continuations.

Repetitions are not identical under the two objectives.
In \ltr{}, repeated excerpts reinforce the same single left-to-right view of each excerpt, and extraction grows logarithmically before saturating.
In \fim{}, the same repeated excerpts appear in different prefix--middle--suffix decompositions.
This makes memorization slower at first, but it can exceed \ltr{} on short-span extraction at high repetition.
Native \fim{} probes further show that while suffixes help, a short true prefix is necessary for extraction.
Replacing the true prefix with a distractor prefix nearly suppresses memorization, while replacing the true suffix with a distractor suffix has a smaller effect.
Our results show that \ltr{} and \fim{} expose different memorization profiles and that memorization in \fim{} remains strongly anchored to the prefix.

\subsection{Limitations and Outlook}
\label{sec:limitations}
Since we pretrain from scratch, we are not able to study frontier-scale models.
Repetition counts are bounded to 128, covering a practically relevant range, but do not allow extrapolation in the limit.
The main conceptual limitation is attribution: under random \fim{} decompositions, a probed span need not match a specific middle span seen during training, so the results do not allow us to trace exact exposures.
Beyond our random-window probes, future work can investigate how prompt position impacts FIM and use span-to-training mappings to test whether the patterns persist across different probes and longer extraction windows.

\clearpage

\section*{Acknowledgements}

We thank Yixuan Xu and Imanol Schlag for their guidance and feedback. This work was supported as part of the Swiss AI Initiative by compute grant infra01 from the Swiss National Supercomputing Centre (CSCS) on Alps.

\section*{Impact Statement}

This work studies how fill-in-the-middle pretraining affects verbatim extraction of repeated text. The results can inform data curation and memorization audits for models with infilling capability. We do not introduce a new attack or release sensitive training examples. The main risk is that better measurement may also help identify settings where extraction is easier; we view this as necessary for evaluating and reducing memorization in deployed systems.

\bibliographystyle{icml2026}
\bibliography{references}

@inproceedings{mmlu,
  author       = {Dan Hendrycks and
                  Collin Burns and
                  Steven Basart and
                  Andy Zou and
                  Mantas Mazeika and
                  Dawn Song and
                  Jacob Steinhardt},
  title        = {Measuring Massive Multitask Language Understanding},
  booktitle    = {9th International Conference on Learning Representations, {ICLR} 2021,
                  Virtual Event, Austria, May 3-7, 2021},
  publisher    = {OpenReview.net},
  year         = {2021},
  url          = {https://openreview.net/forum?id=d7KBjmI3GmQ},
  bibsource    = {dblp computer science bibliography, https://dblp.org}
}

@inproceedings{hellaswag,
  author       = {Rowan Zellers and
                  Ari Holtzman and
                  Yonatan Bisk and
                  Ali Farhadi and
                  Yejin Choi},
  editor       = {Anna Korhonen and
                  David R. Traum and
                  Llu{\'{\i}}s M{\`{a}}rquez},
  title        = {HellaSwag: Can a Machine Really Finish Your Sentence?},
  booktitle    = {Proceedings of the 57th Conference of the Association for Computational
                  Linguistics, {ACL} 2019, Florence, Italy, July 28- August 2, 2019,
                  Volume 1: Long Papers},
  pages        = {4791--4800},
  publisher    = {Association for Computational Linguistics},
  year         = {2019},
  url          = {https://doi.org/10.18653/v1/p19-1472},
  doi          = {10.18653/V1/P19-1472},
  bibsource    = {dblp computer science bibliography, https://dblp.org}
}

@article{arc,
  author       = {Peter Clark and
                  Isaac Cowhey and
                  Oren Etzioni and
                  Tushar Khot and
                  Ashish Sabharwal and
                  Carissa Schoenick and
                  Oyvind Tafjord},
  title        = {Think you have Solved Question Answering? Try ARC, the {AI2} Reasoning
                  Challenge},
  journal      = {CoRR},
  volume       = {abs/1803.05457},
  year         = {2018},
  url          = {http://arxiv.org/abs/1803.05457},
  eprinttype   = {arXiv},
  eprint       = {1803.05457},
  bibsource    = {dblp computer science bibliography, https://dblp.org}
}

@inproceedings{commonsenseqa,
  author       = {Alon Talmor and
                  Jonathan Herzig and
                  Nicholas Lourie and
                  Jonathan Berant},
  editor       = {Jill Burstein and
                  Christy Doran and
                  Thamar Solorio},
  title        = {CommonsenseQA: {A} Question Answering Challenge Targeting Commonsense
                  Knowledge},
  booktitle    = {Proceedings of the 2019 Conference of the North American Chapter of
                  the Association for Computational Linguistics: Human Language Technologies,
                  {NAACL-HLT} 2019, Minneapolis, MN, USA, June 2-7, 2019, Volume 1 (Long
                  and Short Papers)},
  pages        = {4149--4158},
  publisher    = {Association for Computational Linguistics},
  year         = {2019},
  url          = {https://doi.org/10.18653/v1/n19-1421},
  doi          = {10.18653/V1/N19-1421},
  bibsource    = {dblp computer science bibliography, https://dblp.org}
}

@inproceedings{piqa,
  author       = {Yonatan Bisk and
                  Rowan Zellers and
                  Ronan Le Bras and
                  Jianfeng Gao and
                  Yejin Choi},
  title        = {{PIQA:} Reasoning about Physical Commonsense in Natural Language},
  booktitle    = {The Thirty-Fourth {AAAI} Conference on Artificial Intelligence, {AAAI}
                  2020, The Thirty-Second Innovative Applications of Artificial Intelligence
                  Conference, {IAAI} 2020, The Tenth {AAAI} Symposium on Educational
                  Advances in Artificial Intelligence, {EAAI} 2020, New York, NY, USA,
                  February 7-12, 2020},
  pages        = {7432--7439},
  publisher    = {{AAAI} Press},
  year         = {2020},
  url          = {https://doi.org/10.1609/aaai.v34i05.6239},
  doi          = {10.1609/AAAI.V34I05.6239},
  bibsource    = {dblp computer science bibliography, https://dblp.org}
}

@inproceedings{winogrande,
  author       = {Keisuke Sakaguchi and
                  Ronan Le Bras and
                  Chandra Bhagavatula and
                  Yejin Choi},
  title        = {WinoGrande: An Adversarial Winograd Schema Challenge at Scale},
  booktitle    = {The Thirty-Fourth {AAAI} Conference on Artificial Intelligence, {AAAI}
                  2020, The Thirty-Second Innovative Applications of Artificial Intelligence
                  Conference, {IAAI} 2020, The Tenth {AAAI} Symposium on Educational
                  Advances in Artificial Intelligence, {EAAI} 2020, New York, NY, USA,
                  February 7-12, 2020},
  pages        = {8732--8740},
  publisher    = {{AAAI} Press},
  year         = {2020},
  url          = {https://doi.org/10.1609/aaai.v34i05.6399},
  doi          = {10.1609/AAAI.V34I05.6399},
  bibsource    = {dblp computer science bibliography, https://dblp.org}
}

@inproceedings{wikitext,
  author       = {Stephen Merity and
                  Caiming Xiong and
                  James Bradbury and
                  Richard Socher},
  title        = {Pointer Sentinel Mixture Models},
  booktitle    = {5th International Conference on Learning Representations, {ICLR} 2017,
                  Toulon, France, April 24-26, 2017, Conference Track Proceedings},
  publisher    = {OpenReview.net},
  year         = {2017},
  url          = {https://openreview.net/forum?id=Byj72udxe},
  bibsource    = {dblp computer science bibliography, https://dblp.org}
}

@misc{lm-eval-harness,
  author       = {Gao, Leo and Tow, Jonathan and Abbasi, Baber and Biderman, Stella and Black, Sid and DiPofi, Anthony and Foster, Charles and Golding, Laurence and Hsu, Jeffrey and Le Noac'h, Alain and Li, Haonan and McDonell, Kyle and Muennighoff, Niklas and Ociepa, Chris and Phang, Jason and Reynolds, Laria and Schoelkopf, Hailey and Skowron, Aviya and Sutawika, Lintang and Tang, Eric and Thite, Anish and Wang, Ben and Wang, Kevin and Zou, Andy},
  title        = {A framework for few-shot language model evaluation},
  month        = 12,
  year         = 2023,
  publisher    = {Zenodo},
  version      = {v0.4.0},
  doi          = {10.5281/zenodo.10256836},
  url          = {https://zenodo.org/records/10256836}
}

@inproceedings{rouge,
    title = "{ROUGE}: A Package for Automatic Evaluation of Summaries",
    author = "Lin, Chin-Yew",
    booktitle = "Text Summarization Branches Out",
    month = jul,
    year = "2004",
    address = "Barcelona, Spain",
    publisher = "Association for Computational Linguistics",
    url = "https://aclanthology.org/W04-1013/",
    pages = "74--81"
}

@inproceedings{carlini2019secret,
  title={The secret sharer: Evaluating and testing unintended memorization in neural networks},
  author={Carlini, Nicholas and Liu, Chang and Erlingsson, {\'U}lfar and Kos, Jernej and Song, Dawn},
  booktitle={28th USENIX security symposium (USENIX security 19)},
  pages={267--284},
  year={2019}
}

@inproceedings {carlini2021extracting,
author = {Nicholas Carlini and Florian Tram{\`e}r and Eric Wallace and Matthew Jagielski and Ariel Herbert-Voss and Katherine Lee and Adam Roberts and Tom Brown and Dawn Song and {\'U}lfar Erlingsson and Alina Oprea and Colin Raffel},
title = {Extracting Training Data from Large Language Models},
booktitle = {30th USENIX Security Symposium (USENIX Security 21)},
year = {2021},
isbn = {978-1-939133-24-3},
pages = {2633--2650},
url = {https://www.usenix.org/conference/usenixsecurity21/presentation/carlini-extracting},
publisher = {USENIX Association},
month = aug
}

@misc{deepseekai2025deepseekv3technicalreport,
      title={DeepSeek-V3 Technical Report}, 
      author={DeepSeek-AI and Aixin Liu and Bei Feng and Bing Xue and Bingxuan Wang and Bochao Wu and Chengda Lu and Chenggang Zhao and Chengqi Deng and Chenyu Zhang and Chong Ruan and Damai Dai and Daya Guo and Dejian Yang and Deli Chen and Dongjie Ji and Erhang Li and Fangyun Lin and Fucong Dai and Fuli Luo and Guangbo Hao and Guanting Chen and Guowei Li and H. Zhang and Han Bao and Hanwei Xu and Haocheng Wang and Haowei Zhang and Honghui Ding and Huajian Xin and Huazuo Gao and Hui Li and Hui Qu and J. L. Cai and Jian Liang and Jianzhong Guo and Jiaqi Ni and Jiashi Li and Jiawei Wang and Jin Chen and Jingchang Chen and Jingyang Yuan and Junjie Qiu and Junlong Li and Junxiao Song and Kai Dong and Kai Hu and Kaige Gao and Kang Guan and Kexin Huang and Kuai Yu and Lean Wang and Lecong Zhang and Lei Xu and Leyi Xia and Liang Zhao and Litong Wang and Liyue Zhang and Meng Li and Miaojun Wang and Mingchuan Zhang and Minghua Zhang and Minghui Tang and Mingming Li and Ning Tian and Panpan Huang and Peiyi Wang and Peng Zhang and Qiancheng Wang and Qihao Zhu and Qinyu Chen and Qiushi Du and R. J. Chen and R. L. Jin and Ruiqi Ge and Ruisong Zhang and Ruizhe Pan and Runji Wang and Runxin Xu and Ruoyu Zhang and Ruyi Chen and S. S. Li and Shanghao Lu and Shangyan Zhou and Shanhuang Chen and Shaoqing Wu and Shengfeng Ye and Shengfeng Ye and Shirong Ma and Shiyu Wang and Shuang Zhou and Shuiping Yu and Shunfeng Zhou and Shuting Pan and T. Wang and Tao Yun and Tian Pei and Tianyu Sun and W. L. Xiao and Wangding Zeng and Wanjia Zhao and Wei An and Wen Liu and Wenfeng Liang and Wenjun Gao and Wenqin Yu and Wentao Zhang and X. Q. Li and Xiangyue Jin and Xianzu Wang and Xiao Bi and Xiaodong Liu and Xiaohan Wang and Xiaojin Shen and Xiaokang Chen and Xiaokang Zhang and Xiaosha Chen and Xiaotao Nie and Xiaowen Sun and Xiaoxiang Wang and Xin Cheng and Xin Liu and Xin Xie and Xingchao Liu and Xingkai Yu and Xinnan Song and Xinxia Shan and Xinyi Zhou and Xinyu Yang and Xinyuan Li and Xuecheng Su and Xuheng Lin and Y. K. Li and Y. Q. Wang and Y. X. Wei and Y. X. Zhu and Yang Zhang and Yanhong Xu and Yanhong Xu and Yanping Huang and Yao Li and Yao Zhao and Yaofeng Sun and Yaohui Li and Yaohui Wang and Yi Yu and Yi Zheng and Yichao Zhang and Yifan Shi and Yiliang Xiong and Ying He and Ying Tang and Yishi Piao and Yisong Wang and Yixuan Tan and Yiyang Ma and Yiyuan Liu and Yongqiang Guo and Yu Wu and Yuan Ou and Yuchen Zhu and Yuduan Wang and Yue Gong and Yuheng Zou and Yujia He and Yukun Zha and Yunfan Xiong and Yunxian Ma and Yuting Yan and Yuxiang Luo and Yuxiang You and Yuxuan Liu and Yuyang Zhou and Z. F. Wu and Z. Z. Ren and Zehui Ren and Zhangli Sha and Zhe Fu and Zhean Xu and Zhen Huang and Zhen Zhang and Zhenda Xie and Zhengyan Zhang and Zhewen Hao and Zhibin Gou and Zhicheng Ma and Zhigang Yan and Zhihong Shao and Zhipeng Xu and Zhiyu Wu and Zhongyu Zhang and Zhuoshu Li and Zihui Gu and Zijia Zhu and Zijun Liu and Zilin Li and Ziwei Xie and Ziyang Song and Ziyi Gao and Zizheng Pan},
      year={2025},
      eprint={2412.19437},
      archivePrefix={arXiv},
      primaryClass={cs.CL},
      url={https://arxiv.org/abs/2412.19437}, 
}

@inproceedings{
xu2026positional,
title={Positional Fragility in {LLM}s: How Offset Effects Reshape Our Understanding of Memorization Risks},
author={Yixuan Xu and Antoine Bosselut and Imanol Schlag},
booktitle={The Thirty-ninth Annual Conference on Neural Information Processing Systems},
year={2026},
url={https://openreview.net/forum?id=7dBPm5c5ue}
}

@inproceedings{
nasr2025scalable,
title={Scalable Extraction of Training Data from Aligned, Production Language Models},
author={Milad Nasr and Javier Rando and Nicholas Carlini and Jonathan Hayase and Matthew Jagielski and A. Feder Cooper and Daphne Ippolito and Christopher A. Choquette-Choo and Florian Tram{\`e}r and Katherine Lee},
booktitle={The Thirteenth International Conference on Learning Representations},
year={2025},
url={https://openreview.net/forum?id=vjel3nWP2a}
}

@inproceedings{carlini2023quantifying,
  author       = {Nicholas Carlini and
                  Daphne Ippolito and
                  Matthew Jagielski and
                  Katherine Lee and
                  Florian Tram{\`{e}}r and
                  Chiyuan Zhang},
  title        = {Quantifying Memorization Across Neural Language Models},
  booktitle    = {The Eleventh International Conference on Learning Representations,
                  {ICLR} 2023, Kigali, Rwanda, May 1-5, 2023},
  publisher    = {OpenReview.net},
  year         = {2023},
  url          = {https://openreview.net/forum?id=TatRHT\_1cK},
  bibsource    = {dblp computer science bibliography, https://dblp.org}
}

@misc{cooper2025extracting,
      title={Extracting memorized pieces of (copyrighted) books from open-weight language models}, 
      author={A. Feder Cooper and Mark A. Lemley and Allison Casasola and Ahmed Ahmed and Aaron Gokaslan and Amy B. Cyphert and Christopher De Sa and Daniel E. Ho and Percy Liang},
      year={2026},
      eprint={2505.12546},
      archivePrefix={arXiv},
      primaryClass={cs.CL},
      url={https://arxiv.org/abs/2505.12546}, 
}

@inproceedings{hayes2025measuring,
  title={Measuring memorization in language models via probabilistic extraction},
  author={Hayes, Jamie and Swanberg, Marika and Chaudhari, Harsh and Yona, Itay and Shumailov, Ilia and Nasr, Milad and Choquette-Choo, Christopher A and Lee, Katherine and Cooper, A Feder},
  booktitle={Proceedings of the 2025 Conference of the Nations of the Americas Chapter of the Association for Computational Linguistics: Human Language Technologies (Volume 1: Long Papers)},
  pages={9266--9291},
  year={2025}
}

@inproceedings{mattern2023membership,
    title = "Membership Inference Attacks against Language Models via Neighbourhood Comparison",
    author = {Mattern, Justus  and
      Mireshghallah, Fatemehsadat  and
      Jin, Zhijing  and
      Sch{\"o}lkopf, Bernhard  and
      Sachan, Mrinmaya  and
      Berg-Kirkpatrick, Taylor},
    editor = "Rogers, Anna  and
      Boyd-Graber, Jordan  and
      Okazaki, Naoaki",
    booktitle = "Findings of the Association for Computational Linguistics: ACL 2023",
    month = jul,
    year = "2023",
    address = "Toronto, Canada",
    publisher = "Association for Computational Linguistics",
    url = "https://aclanthology.org/2023.findings-acl.719/",
    doi = "10.18653/v1/2023.findings-acl.719",
    pages = "11330--11343"
}

@inproceedings{
shi2024detecting,
title={Detecting Pretraining Data from Large Language Models},
author={Weijia Shi and Anirudh Ajith and Mengzhou Xia and Yangsibo Huang and Daogao Liu and Terra Blevins and Danqi Chen and Luke Zettlemoyer},
booktitle={The Twelfth International Conference on Learning Representations},
year={2024},
url={https://openreview.net/forum?id=zWqr3MQuNs}
}

@inproceedings{lee2022deduplicating,
    title = "Deduplicating Training Data Makes Language Models Better",
    author = "Lee, Katherine  and
      Ippolito, Daphne  and
      Nystrom, Andrew  and
      Zhang, Chiyuan  and
      Eck, Douglas  and
      Callison-Burch, Chris  and
      Carlini, Nicholas",
    editor = "Muresan, Smaranda  and
      Nakov, Preslav  and
      Villavicencio, Aline",
    booktitle = "Proceedings of the 60th Annual Meeting of the Association for Computational Linguistics (Volume 1: Long Papers)",
    month = may,
    year = "2022",
    address = "Dublin, Ireland",
    publisher = "Association for Computational Linguistics",
    url = "https://aclanthology.org/2022.acl-long.577/",
    doi = "10.18653/v1/2022.acl-long.577",
    pages = "8424--8445"
}

@inproceedings{kandpal2022deduplicating,
  author       = {Nikhil Kandpal and
                  Eric Wallace and
                  Colin Raffel},
  editor       = {Kamalika Chaudhuri and
                  Stefanie Jegelka and
                  Le Song and
                  Csaba Szepesv{\'{a}}ri and
                  Gang Niu and
                  Sivan Sabato},
  title        = {Deduplicating Training Data Mitigates Privacy Risks in Language Models},
  booktitle    = {International Conference on Machine Learning, {ICML} 2022, 17-23 July
                  2022, Baltimore, Maryland, {USA}},
  series       = {Proceedings of Machine Learning Research},
  pages        = {10697--10707},
  publisher    = {{PMLR}},
  year         = {2022},
  url          = {https://proceedings.mlr.press/v162/kandpal22a.html},
  bibsource    = {dblp computer science bibliography, https://dblp.org}
}

@inproceedings{huang2024demystifying,
    title = "Demystifying Verbatim Memorization in Large Language Models",
    author = "Huang, Jing  and
      Yang, Diyi  and
      Potts, Christopher",
    editor = "Al-Onaizan, Yaser  and
      Bansal, Mohit  and
      Chen, Yun-Nung",
    booktitle = "Proceedings of the 2024 Conference on Empirical Methods in Natural Language Processing",
    month = nov,
    year = "2024",
    address = "Miami, Florida, USA",
    publisher = "Association for Computational Linguistics",
    url = "https://aclanthology.org/2024.emnlp-main.598/",
    doi = "10.18653/v1/2024.emnlp-main.598",
    pages = "10711--10732"
}

@article{kharitonov2022bpe,
  author       = {Eugene Kharitonov and
                  Marco Baroni and
                  Dieuwke Hupkes},
  title        = {How {BPE} Affects Memorization in Transformers},
  journal      = {CoRR},
  volume       = {abs/2110.02782},
  year         = {2021},
  url          = {https://arxiv.org/abs/2110.02782},
  eprinttype   = {arXiv},
  eprint       = {2110.02782},
  bibsource    = {dblp computer science bibliography, https://dblp.org}
}

@inproceedings{
zhang2023counterfactual,
title={Counterfactual Memorization in Neural Language Models},
author={Chiyuan Zhang and Daphne Ippolito and Katherine Lee and Matthew Jagielski and Florian Tram{\`e}r and Nicholas Carlini},
booktitle={Thirty-seventh Conference on Neural Information Processing Systems},
year={2023},
url={https://openreview.net/forum?id=67o9UQgTD0}
}

@article{shilov2026mosaic,
  author = {Shilov, Igor and Meeus, Matthieu and de Montjoye, Yves-Alexandre},
  title = {The mosaic memory of large language models},
  journal = {Nature Communications},
  year = {2026},
  month = {Jan},
  volume = {17},
  number = {1},
  publisher = {Springer Science and Business Media LLC},
  doi = {10.1038/s41467-026-68603-0},
  url = {http://dx.doi.org/10.1038/s41467-026-68603-0},
  issn = {2041-1723}
}

@article{liu2024lost,
    title = "Lost in the Middle: How Language Models Use Long Contexts",
    author = "Liu, Nelson F.  and
      Lin, Kevin  and
      Hewitt, John  and
      Paranjape, Ashwin  and
      Bevilacqua, Michele  and
      Petroni, Fabio  and
      Liang, Percy",
    journal = "Transactions of the Association for Computational Linguistics",
    volume = "12",
    year = "2024",
    address = "Cambridge, MA",
    publisher = "MIT Press",
    url = "https://aclanthology.org/2024.tacl-1.9/",
    doi = "10.1162/tacl_a_00638",
    pages = "157--173"
}

@article{bavarian2022efficient,
  title={Efficient Training of Language Models to Fill in the Middle},
  author={Mo Bavarian and Heewoo Jun and Nikolas A. Tezak and John Schulman and Christine McLeavey and Jerry Tworek and Mark Chen},
  journal={ArXiv},
  year={2022},
  volume={abs/2207.14255},
  url={https://api.semanticscholar.org/CorpusID:251135268}
}

@inproceedings{
fried2023incoder,
title={InCoder: A Generative Model for Code Infilling and Synthesis},
author={Daniel Fried and Armen Aghajanyan and Jessy Lin and Sida Wang and Eric Wallace and Freda Shi and Ruiqi Zhong and Scott Yih and Luke Zettlemoyer and Mike Lewis},
booktitle={The Eleventh International Conference on Learning Representations },
year={2023},
url={https://openreview.net/forum?id=hQwb-lbM6EL}
}

@article{
li2023starcoder,
title={StarCoder: may the source be with you!},
author={Raymond Li and Loubna Ben allal and Yangtian Zi and Niklas Muennighoff and Denis Kocetkov and Chenghao Mou and Marc Marone and Christopher Akiki and Jia LI and Jenny Chim and Qian Liu and Evgenii Zheltonozhskii and Terry Yue Zhuo and Thomas Wang and Olivier Dehaene and Joel Lamy-Poirier and Joao Monteiro and Nicolas Gontier and Ming-Ho Yee and Logesh Kumar Umapathi and Jian Zhu and Ben Lipkin and Muhtasham Oblokulov and Zhiruo Wang and Rudra Murthy and Jason T Stillerman and Siva Sankalp Patel and Dmitry Abulkhanov and Marco Zocca and Manan Dey and Zhihan Zhang and Urvashi Bhattacharyya and Wenhao Yu and Sasha Luccioni and Paulo Villegas and Fedor Zhdanov and Tony Lee and Nadav Timor and Jennifer Ding and Claire S Schlesinger and Hailey Schoelkopf and Jan Ebert and Tri Dao and Mayank Mishra and Alex Gu and Carolyn Jane Anderson and Brendan Dolan-Gavitt and Danish Contractor and Siva Reddy and Daniel Fried and Dzmitry Bahdanau and Yacine Jernite and Carlos Mu{\~n}oz Ferrandis and Sean Hughes and Thomas Wolf and Arjun Guha and Leandro Von Werra and Harm de Vries},
journal={Transactions on Machine Learning Research},
issn={2835-8856},
year={2023},
url={https://openreview.net/forum?id=KoFOg41haE},
note={Reproducibility Certification}
}

@article{roziere2023code,
  title={Code Llama: Open Foundation Models for Code},
  author={Baptiste Rozi{\`e}re and Jonas Gehring and Fabian Gloeckle and Sten Sootla and Itai Gat and Xiaoqing Tan and Yossi Adi and Jingyu Liu and Tal Remez and J{\'e}r{\'e}my Rapin and Artyom Kozhevnikov and I. Evtimov and Joanna Bitton and Manish P Bhatt and Cristian Canton Ferrer and Aaron Grattafiori and Wenhan Xiong and Alexandre D'efossez and Jade Copet and Faisal Azhar and Hugo Touvron and Louis Martin and Nicolas Usunier and Thomas Scialom and Gabriel Synnaeve},
  journal={ArXiv},
  year={2023},
  volume={abs/2308.12950},
  url={https://api.semanticscholar.org/CorpusID:261100919}
}

@inproceedings{
penedo2024fineweb,
title={The FineWeb Datasets: Decanting the Web for the Finest Text Data at Scale},
author={Guilherme Penedo and Hynek Kydl{\'\i}{\v{c}}ek and Loubna Ben allal and Anton Lozhkov and Margaret Mitchell and Colin Raffel and Leandro Von Werra and Thomas Wolf},
booktitle={The Thirty-eight Conference on Neural Information Processing Systems Datasets and Benchmarks Track},
year={2024},
url={https://openreview.net/forum?id=n6SCkn2QaG}
}

@misc{gutenberg,
  author       = {{Project Gutenberg}},
  title        = {Project Gutenberg},
  year         = {n.d.},
  howpublished = {\url{https://www.gutenberg.org}},
  note         = {Accessed: 2026-05-04}
}

@inproceedings{
chen2025parapo,
title={Para{PO}: Aligning Language Models to Reduce Verbatim Reproduction of Pre-training Data},
author={Tong Chen and Faeze Brahman and Jiacheng Liu and Niloofar Mireshghallah and Weijia Shi and Pang Wei Koh and Luke Zettlemoyer and Hannaneh Hajishirzi},
booktitle={Second Conference on Language Modeling},
year={2025},
url={https://openreview.net/forum?id=Uic3ojVhXh}
}

@article{grattafiori2024llama3herdmodels,
  author       = {{Llama Team}},
  title        = {The Llama 3 Herd of Models},
  journal      = {CoRR},
  volume       = {abs/2407.21783},
  year         = {2024},
  url          = {https://doi.org/10.48550/arXiv.2407.21783},
  doi          = {10.48550/ARXIV.2407.21783},
  eprinttype   = {arXiv},
  eprint       = {2407.21783},
  bibsource    = {dblp computer science bibliography, https://dblp.org}
}

\newpage
\appendix
\crefalias{section}{appendix}
\onecolumn
\section{Experimental Details}
\label{app:experimental-details}

\subsection{Training Parameters}
\label{app:parameters}
Both paired models use the Llama 3.2 3B architecture implemented in Megatron-LM with packed sequences and FlashAttention. \Cref{tab:llama-config} gives the fixed backbone configuration.

\begin{table}[H]
  \centering
  \caption{\textbf{Backbone parameters.} The \ltr{} and \fim{} runs use the same tokenizer, architecture, precision, and context-length configuration.}
  \label{tab:llama-config}
  \begin{tabular}{lr}
    \toprule
    Parameter & Value \\
    \midrule
    Layers & 28 \\
    Hidden size & 3072 \\
    Attention heads & 24 \\
    KV heads & 8 \\
    FFN hidden size & 8192 \\
    Vocab size & 128256 \\
    Max position embeddings & 131072 \\
    RoPE base & 500000 \\
    Precision & bfloat16 \\
    Dropout & 0 \\
    \bottomrule
  \end{tabular}
\end{table}

Both paired runs use packed THD-format sequences, sequence length 16{,}384, micro-batch size 1, no dropout, and global batch size 2048 across 64 GH200 GPUs. One optimization step therefore consumes 33{,}554{,}432 tokens. The \ltr{} run performs 3{,}057 updates over 102.58B tokens. The \fim{} run performs 3{,}064 updates over 102.81B tokens. We release training logs at \url{https://wandb.ai/memorization-study-fim-team/memorization-study-fim/} and model checkpoints on HuggingFace: \href{https://huggingface.co/tvonarx/memfim-fim-3b}{FIM 3B}, \href{https://huggingface.co/tvonarx/memfim-ltr-3b}{LTR 3B}, \href{https://huggingface.co/tvonarx/memfim-fim-1b}{FIM 1B}, \href{https://huggingface.co/tvonarx/memfim-ltr-1b}{LTR 1B}.

\subsection{FIM Formatting}
\label{app:fim-formatting}

For each \fim{} document, following \citet{bavarian2022efficient}, we randomly sample two split points within the document segment, yielding a prefix \(\colorbox{blue!12}{\(\mathbf{P}\)}\), middle span \(\colorbox{red!12}{\(\mathbf{M}\)}\), and suffix \(\colorbox{orange!18}{\(\mathbf{S}\)}\). The \fim{} condition reuses reserved Llama special-token IDs, so we do not resize the embedding table.

\[
\texttt{<|fim\_prefix|>}=128002,\qquad
\texttt{<|fim\_middle|>}=128003,\qquad
\texttt{<|fim\_suffix|>}=128005.
\]

The \ltr{} format keeps the original order:
\[
{\small
\colorbox{blue!12}{\(\mathbf{P}\)}
\colorbox{red!12}{\(\mathbf{M}\)}
\colorbox{orange!18}{\(\mathbf{S}\)}
\texttt{<|eos\_token|>}
}
\]

The \fim{} format moves the middle span after its surrounding context:
\[
{\small
\texttt{<|fim\_prefix|>} \colorbox{blue!12}{\(\mathbf{P}\)}
\texttt{<|fim\_suffix|>} \colorbox{orange!18}{\(\mathbf{S}\)}
\texttt{<|fim\_middle|>} \colorbox{red!12}{\(\mathbf{M}\)}
\texttt{<|eos\_token|>}
}
\]

For FineWeb, the \fim{} training uses a 50\% \fim{} / 50\% \ltr{} mixture. For Gutenberg, every 4096-token excerpt is formatted using the \fim{}-format. The \ltr{}-model is only trained on \ltr{} sequences and contains no \fim{} sentinels.

\subsection{Gutenberg Filtering and Deduplication}
\label{app:gutenberg-filter}

We filter Project Gutenberg to obtain fixed 4096-token excerpts whose later extraction is due to controlled exposure rather than prior web familiarity. Starting from the English split of Project Gutenberg on HuggingFace\footnote{\texttt{manu/project\_gutenberg}}, we strip standard Gutenberg headers, footers, licenses, and archive boilerplate. From each cleaned book, we keep characters 10{,}000--80{,}000, tokenize with the Llama 3.2 tokenizer, split into non-overlapping 4096-token windows, score each window with the FineWeb-only Llama 3.2 3B checkpoint, and keep the highest-PPL window per book. We then remove windows with PPL \(>500\), which were mostly indices, glossary fragments, OCR artifacts, or unusual formatting.

We deduplicate with both semantic and lexical evidence. Excerpts are embedded with \texttt{nomic-ai/nomic-embed-text-v1.5}; a pair is removed only if cosine similarity is at least 0.96 and token 5-gram Jaccard overlap is at least 0.20. For each duplicate cluster, we keep the highest-PPL excerpt. This reduces 128{,}003 scored windows to 33{,}720 final excerpts.

The final schedule has 12 repetition buckets with exposures \(1,2,3,4,8,16,24,32,48,64,96,128\). Each bucket contains 2{,}810 base excerpts, for 1{,}197{,}060 Gutenberg training documents after replication. Bucket assignment is balanced by FineWeb-checkpoint PPL; bucket means range from 36.895227 to 36.895758. The \ltr{}-format Gutenberg corpus has 4{,}904{,}354{,}820 tokens, and the \fim{}-format Gutenberg corpus has 4{,}907{,}946{,}000 tokens, with the difference coming from \fim{} sentinel tokens.

\section{Additional Experimental Results}
\label{app:additional-experimental-results}

\subsection{Downstream Performance}
\label{app:evaluation-results}

We report the detailed metrics of our matched Llama 3.2 3B models for the LM Evaluation Harness suite \citep{lm-eval-harness} in \cref{tab:downstream-control-by-scale}. The scores of the 1B-scale ablation in \cref{app:ablation-model-size} are also reported.

\begin{table}[H]
  \centering
  \caption{\textbf{Downstream quality-control suite by model scale.} Accuracy tasks are reported in \%; higher is better. Lower is better for Wikitext word perplexity. Within each scale, \colorbox{green!15}{green} marks the better of \ltr{} and \fim{}, and \colorbox{red!12}{red} marks the worse. Bold marks the best score in the row. \(\Delta\) is \fim{} minus \ltr{} within each scale (pp for accuracy, absolute for PPL).}
  \label{tab:downstream-control-by-scale}
  \begin{tabular}{llcccccc}
    \toprule
    & & \multicolumn{3}{c}{1B} & \multicolumn{3}{c}{3B} \\
    \cmidrule(lr){3-5} \cmidrule(lr){6-8}
    Task & Metric & \ltr{} & \fim{} & \(\Delta\) & \ltr{} & \fim{} & \(\Delta\) \\
    \midrule
    MMLU \citep{mmlu} & acc\,\% \(\uparrow\) &
    \colorbox{green!15}{23.57} & \colorbox{red!12}{22.89} & -0.68 &
    \colorbox{red!12}{23.81} & \colorbox{green!15}{\textbf{24.42}} & +0.61 \\
    HellaSwag \citep{hellaswag} & acc\,\% \(\uparrow\) &
    \colorbox{red!12}{32.40} & \colorbox{green!15}{32.65} & +0.25 &
    \colorbox{red!12}{36.36} & \colorbox{green!15}{\textbf{36.95}} & +0.60 \\
    ARC-Challenge \citep{arc} & acc\,\% \(\uparrow\) &
    19.28 & 19.28 & +0.00 &
    \colorbox{green!15}{\textbf{22.18}} & \colorbox{red!12}{21.25} & -0.94 \\
    ARC-Easy \citep{arc} & acc\,\% \(\uparrow\) &
    \colorbox{green!15}{48.36} & \colorbox{red!12}{47.52} & -0.84 &
    \colorbox{green!15}{\textbf{53.24}} & \colorbox{red!12}{52.53} & -0.72 \\
    CommonsenseQA \citep{commonsenseqa} & acc\,\% \(\uparrow\) &
    19.57 & 19.57 & +0.00 &
    \colorbox{green!15}{\textbf{19.98}} & \colorbox{red!12}{19.74} & -0.25 \\
    PIQA \citep{piqa} & acc\,\% \(\uparrow\) &
    \colorbox{green!15}{67.90} & \colorbox{red!12}{67.03} & -0.87 &
    \colorbox{red!12}{69.80} & \colorbox{green!15}{\textbf{70.40}} & +0.60 \\
    WinoGrande \citep{winogrande} & acc\,\% \(\uparrow\) &
    \colorbox{green!15}{\textbf{51.22}} & \colorbox{red!12}{51.07} & -0.16 &
    \colorbox{green!15}{51.14} & \colorbox{red!12}{50.43} & -0.71 \\
    Wikitext \citep{wikitext} & PPL\hfill\(\downarrow\) &
    \colorbox{green!15}{26.19} & \colorbox{red!12}{26.30} & +0.11 &
    \colorbox{green!15}{\textbf{25.25}} & \colorbox{red!12}{26.98} & +1.73 \\
    \bottomrule
  \end{tabular}
\end{table}

\subsection{Model size ablation}
\label{app:ablation-model-size}

Memorization has been shown to increase with model capacity \citep{carlini2023quantifying}. To test the validity and generalizability of our conclusions, we train paired Llama 3.2 1B models in the same conditions.

We observe that, as expected, both downstream performance and verbatim memorization decrease at smaller scale.
\cref{tab:downstream-control-by-scale} reports the downstream comparison, and \cref{fig:ltr-window-position-rouge-1b-3b} shows reduced memorization relative to the 3B variant.
Because exact extraction on the random-window probes used in \cref{sec:prefix-only-extraction} is too rare at 1B scale for a stable comparison, we focus on ROUGE-L instead.

Importantly, note that the relative trends between \ltr{} and \fim{} remain consistent with our main results in \cref{sec:prefix-only-extraction}.

\subsection{Additional Figures}
\label{app:additional-figures}

\begin{figure}[h]
  \centering
  \includegraphics[width=0.48\linewidth]{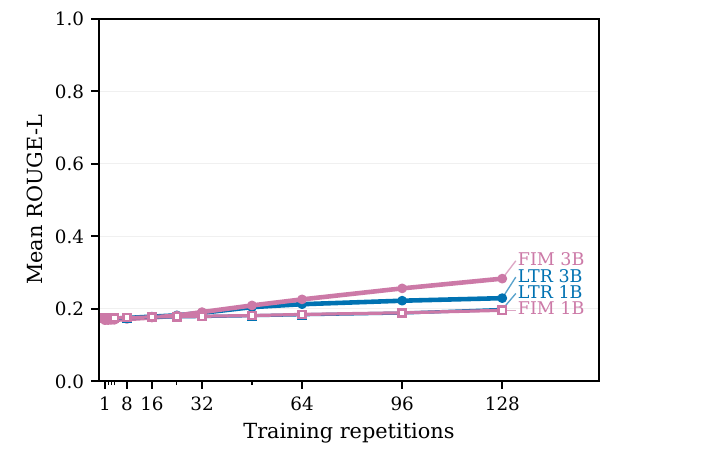}
  \hfill
  \includegraphics[width=0.48\linewidth]{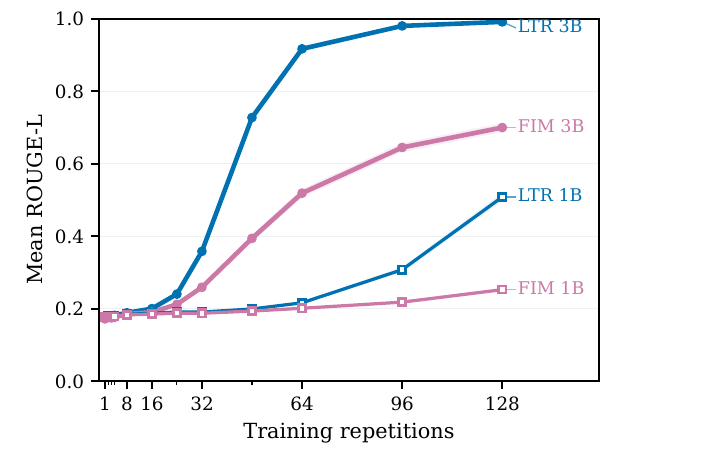}
  \caption{
  \textbf{Mean ROUGE-L under prefix probing for 1B and 3B models}, evaluated on 10 uniformly sampled windows per excerpt (left) and on the first window of each excerpt (right).
  Each prompt uses 100 prefix tokens to generate a 32-token continuation.
  Filled circles denote 3B models; hollow squares denote 1B models.
  The large gap between first-window and uniformly sampled-window probing indicates that recall is anchored near the beginning of repeated excerpts, consistent with positional fragility observed by \citet{xu2026positional}.
  }
  \label{fig:ltr-window-position-rouge-1b-3b}
\end{figure}

\begin{figure}[h]
  \centering
  \begin{subfigure}{0.44\linewidth}
    \centering
    \includegraphics[width=\linewidth]{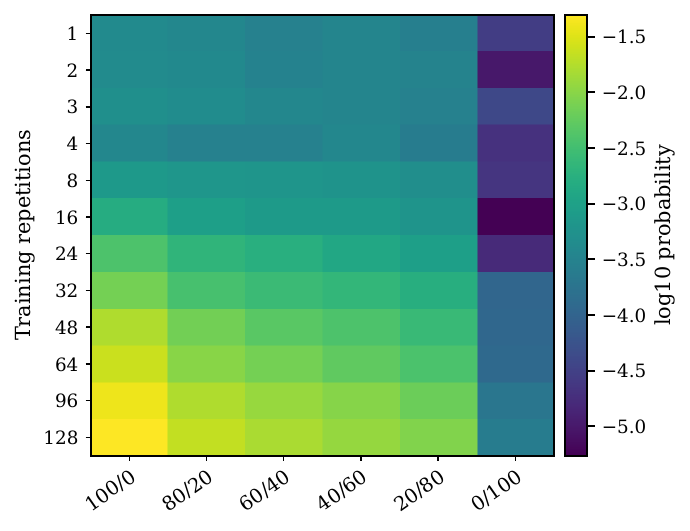}
    \caption{Mean per-token Cooper probability.}
  \end{subfigure}
  \hfill
  \begin{subfigure}{0.44\linewidth}
    \centering
    \includegraphics[width=\linewidth]{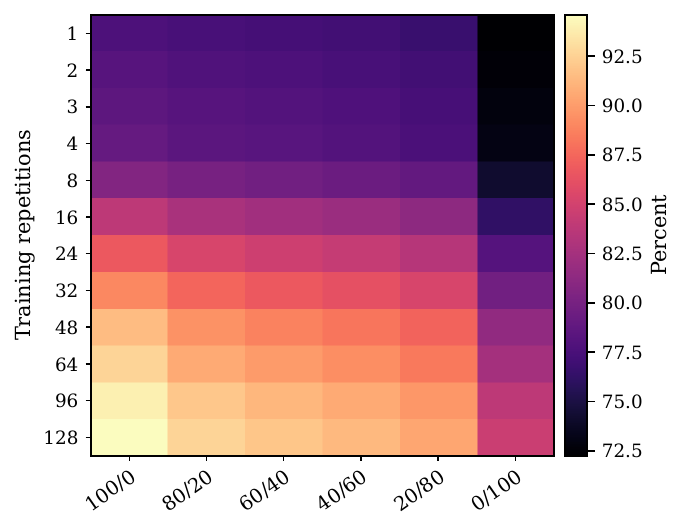}
    \caption{Target tokens in top-$k$.}
  \end{subfigure}

  \caption{\textbf{Native \fim{} geometry by repetition bucket.} Heatmaps separate the prefix--suffix effect across repetition levels. The x-axis varies \texttt{prefix}/\texttt{suffix} lengths.
}
  \label{fig:native-fim-geometry-heatmaps}
\end{figure}

\begin{figure}[h]
  \centering
  \begin{subfigure}{0.44\linewidth}
    \centering
    \includegraphics[width=\linewidth]{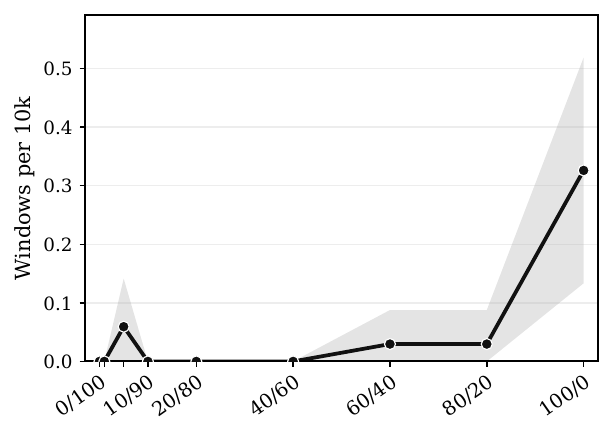}
    \caption{Extractability}
  \end{subfigure}
  \hfill
  \begin{subfigure}{0.44\linewidth}
    \centering
    \includegraphics[width=\linewidth]{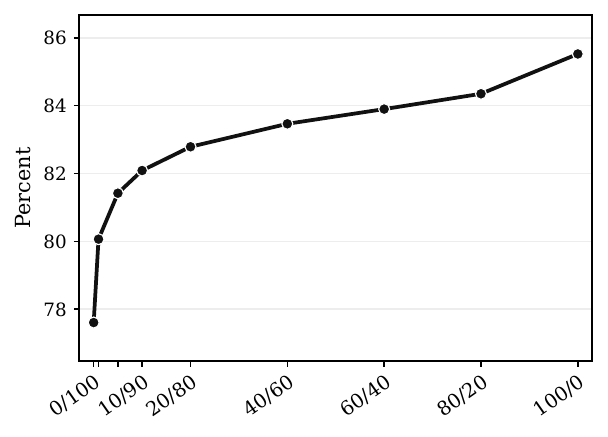}
    \caption{Target tokens in top-$k$}
  \end{subfigure}

  \vspace{0.3em}

  \begin{subfigure}{0.44\linewidth}
    \centering
    \includegraphics[width=\linewidth]{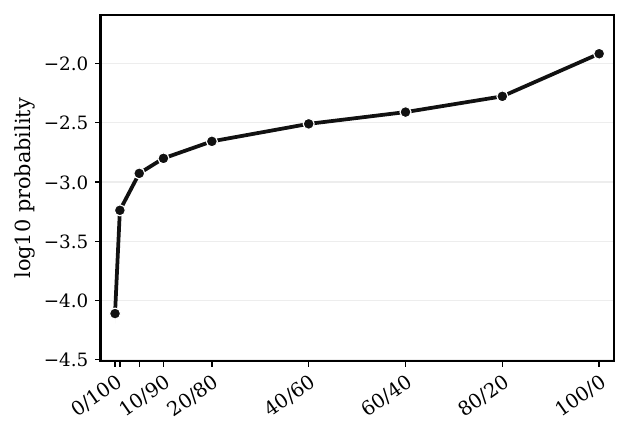}
    \caption{Mean per-token top-$k$ renormalized log-probability.}
  \end{subfigure}
  \hfill
  \begin{subfigure}{0.44\linewidth}
    \centering
    \includegraphics[width=\linewidth]{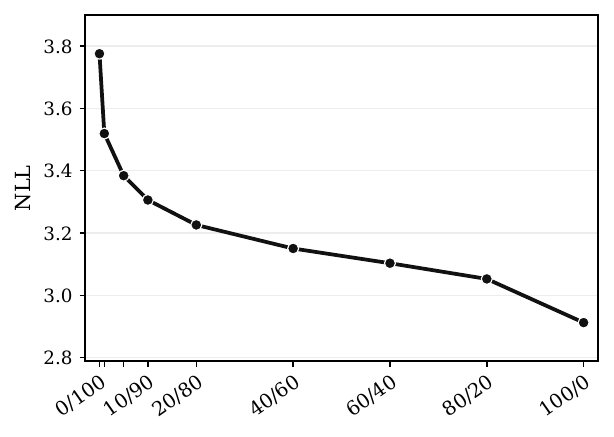}
    \caption{Teacher-forced target NLL}
  \end{subfigure}

  \caption{\textbf{Native \fim{} probing across prefix--suffix geometry.} Metrics are over \textit{all} repetition buckets. The x-axis varies \texttt{prefix}/\texttt{suffix} lengths. Shaded bands are nominal 95\% confidence intervals.}
  \label{fig:native-fim-geometry-profile}
\end{figure}

\cref{fig:ltr-window-position-rouge-1b-3b,fig:native-fim-geometry-heatmaps,fig:native-fim-geometry-profile} show additional figures omitted from the main text.

\subsection{Qualitative Assessment of Memorization}
\label{app:qualitative-examples}

\begin{figure}[h]
    \centering
    \begin{subfigure}[t]{0.48\linewidth}
        \centering
        \includegraphics[width=\linewidth]{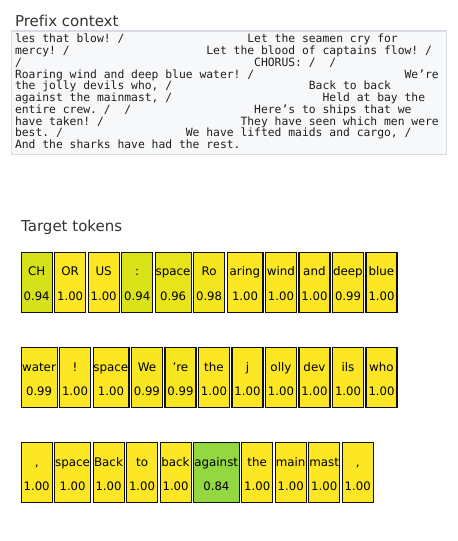}
        \caption{\ltr{}}
    \end{subfigure}
    \hfill
    \begin{subfigure}[t]{0.48\linewidth}
        \centering
        \includegraphics[width=\linewidth]{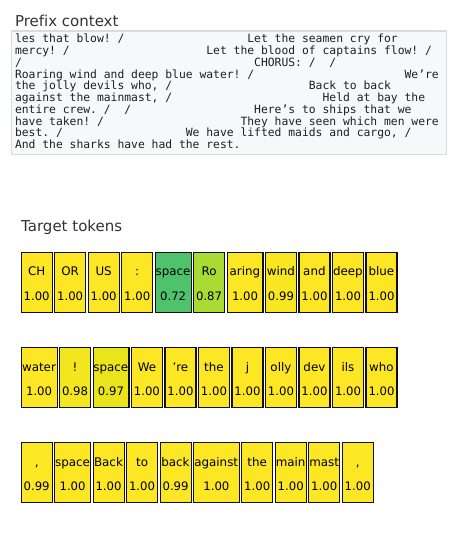}
        \caption{\fim{}}
    \end{subfigure}
    \caption{
    Window that was extracted by both models. Numbers indicate the top-$k$ re-normalized logits of the displayed true target tokens. Repetition 128; source book 54068-0; excerpt 54068-0::window\_0000; target start 100; prefix length 100 tokens; target length 32 tokens; \(\pz{}\) values: \ltr{}=0.711046, \fim{}=0.585069.
    }
    \label{fig:example_both}
\end{figure}

\begin{figure}[h]
    \centering
    \begin{subfigure}[t]{0.48\linewidth}
        \centering
        \includegraphics[width=\linewidth]{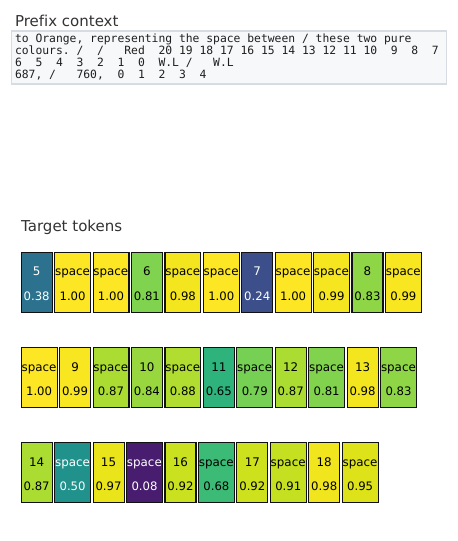}
        \caption{\ltr{}}
    \end{subfigure}
    \hfill
    \begin{subfigure}[t]{0.48\linewidth}
        \centering
        \includegraphics[width=\linewidth]{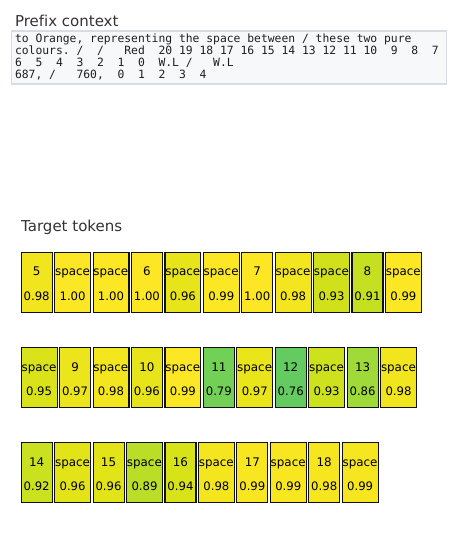}
        \caption{\fim{}}
    \end{subfigure}
    \caption{
    Window that was only extracted by the \fim{}-model. Numbers indicate the top-$k$ re-normalized logits of the displayed true target tokens. Repetition 128; source book 57335-0; excerpt 57335-0::window\_0002; target start 100; prefix length 100 tokens; target length 32 tokens; \(\pz{}\) values: \ltr{}=0.000219776, \fim{}=0.204912.
    }
    \label{fig:example_fim_only}
\end{figure}

\begin{figure}[h]
    \centering
    \begin{subfigure}[t]{0.48\linewidth}
        \centering
        \includegraphics[width=\linewidth, trim={0 1cm 0 0}, clip]{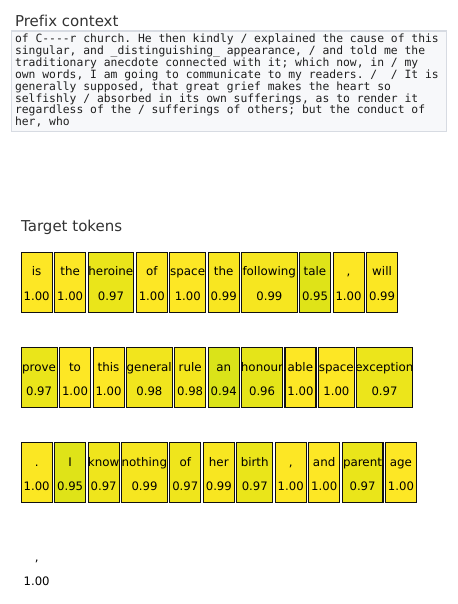}
        \caption{\ltr{}}
    \end{subfigure}
    \hfill
    \begin{subfigure}[t]{0.48\linewidth}
        \centering
        \includegraphics[width=\linewidth, trim={0 1cm 0 0}, clip]{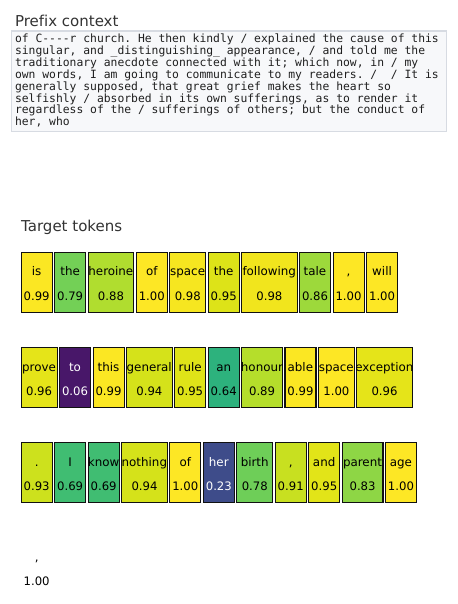}
        \caption{\fim{}}
    \end{subfigure}
    \caption{
    Window that was only extracted by the \ltr{}-model. Numbers indicate the top-$k$ re-normalized logits of the displayed true target tokens. Repetition 128; source book 11326-8; excerpt 11326-8::window\_0003; target start 100; prefix length 100 tokens; target length 32 tokens; \(\pz{}\) values: \ltr{}=0.588202, \fim{}=0.00063823.
    }
    \label{fig:example_ltr_only}
\end{figure}

\cref{fig:example_both,fig:example_fim_only,fig:example_ltr_only} show examples of memorized windows that are extractable by both models (\cref{fig:example_both}), only extractable by the \fim{}-model (\cref{fig:example_fim_only}), and only extractable by the \ltr{}-model (\cref{fig:example_ltr_only}).

\clearpage

\end{document}